\documentclass{article}

\usepackage{PRIMEarxiv}

\usepackage[utf8]{inputenc} % allow utf-8 input
\usepackage[T1]{fontenc}    % use 8-bit T1 fonts
\usepackage{hyperref}       % hyperlinks
\usepackage{url}            % simple URL typesetting
\usepackage{booktabs}       % professional-quality tables
\usepackage{amsfonts}       % blackboard math symbols
\usepackage{nicefrac}       % compact symbols for 1/2, etc.
\usepackage{microtype}      % microtypography
\usepackage{lipsum}
\usepackage{fancyhdr}       % header
\usepackage{graphicx}       % graphics
\usepackage{amsmath} 
\graphicspath{{media/}}     % organize your images and other figures under media/ folder

\title{DL-KDD: Dual-Light KnowleDge Distillation for Action Recognition in the Dark
}

\author{
  Chi-Jui Chang\(^1\)\\
  Institute of Computer Science and Engineering\\
  National Yang Ming Chiao Tung University \\
  Hsinchu\\
  \texttt{jerryyyyy708.cs12@nycu.edu.tw} \\
  \And
  Oscar Tai-Yuan Chen\(^2\) \\
  Institute of Computer Science and Engineering\\
  National Yang Ming Chiao Tung University \\
  Hsinchu\\
  \And
  Vincent S. Tseng\(^{*}\) \\
  Department of Computer Science \\
  National Yang Ming Chiao Tung University, Hsinchu, Taiwan, R.O.C \\
  orcid=0000-0002-4853-1594\\
  *Correspondence: \texttt{vtseng@cs.nycu.edu.tw} \\
}

\begin{document}
\maketitle

\begin{abstract}
Human action recognition in dark videos is a challenging task for computer vision. Recent research focuses on applying dark enhancement methods to improve the visibility of the video. However, such video processing results in the loss of critical information in the original (un-enhanced) video. Conversely, traditional two-stream methods are capable of learning information from both original and processed videos, but it can lead to a significant increase in the computational cost during the inference phase in the task of video classification. To address these challenges, we propose a novel teacher-student video classification framework, named \textit{Dual-Light KnowleDge Distillation for Action Recognition in the Dark (DL-KDD)}. This framework enables the model to learn from both original and enhanced video without introducing additional computational cost during inference. Specifically, DL-KDD utilizes the strategy of knowledge distillation during training. The teacher model is trained with enhanced video, and the student model is trained with both the original video and the soft target generated by the teacher model. This teacher-student framework allows the student model to predict action using only the original input video during inference. In our experiments, the proposed DL-KDD framework outperforms state-of-the-art methods on the ARID, ARID V1.5, and Dark-48 datasets. We achieve the best performance on each dataset and up to a 4.18\% improvement on Dark-48, using only original video inputs, thus avoiding the use of two-stream framework or enhancement modules for inference. We further validate the effectiveness of the distillation strategy in ablative experiments. The results highlight the advantages of our knowledge distillation framework in dark human action recognition.
\end{abstract}

% keywords can be removed

\section{Introduction}
Action Recognition is a popular task in computer vision that can be applied in various real-world applications. For example, surveillance systems \cite{surv} and autonomous vehicles \cite{auto1,auto2}. In recent years, there has been increasing research focusing on this task \cite{slowfast, 21d,3dres,times,swin}.
Compared to action recognition under well-lighted conditions, recognizing action in dark environments is more challenging due to the degradation of the information in videos. In response to this challenge, recent studies \cite{darklight,gcn,dtcm} have proposed various frameworks to achieve better performance with dark video inputs. Common approaches include utilizing light enhancement methods such as ZeroDCE \cite{zerodce} and Gamma Intensity Correction(GIC) to improve the video feature and visibility, followed by 3D convolutional networks like R(2+1)D \cite{21d} or 3D-ResNext \cite{3dres} as the backbone classifiers. Two main architectures to incorporate these components are: 1) directly integrating two models \cite{gcn,dtcm}, 2) using a two-stream method \cite{21d} to improve the accuracy of action prediction from dark videos.

Recent research \cite{gcn, dtcm} focuses on applying enhancements and taking enhanced video as model inputs. While such approaches improve the features contained in videos, the enhancement process often leads to losing original content, which can contain critical information for action recognition.
On the other hand, existing methods \cite{darklight} in dark human action recognition that considers the importance of the original input applied traditional two-stream \cite{ts, i3d} method, which takes both the original and enhanced video as the inputs to the model. This approach significantly increases the computational load, making it slower to perform predictions during inference.
In contrast, using only the original video as input results in a performance gap compared to previously mentioned techniques since the model can get less information from the raw video without enhancement. In summary, the three main challenges for current research on dark human action recognition are

\textbf{Information Completeness:} Ensuring the model to learn from enhanced video without losing essential information from the original video.

\textbf{Complexity Tradeoff:} Making full use of original video and consider the enhanced feature without additional model complexity.

\textbf{Consistent Performance:} Improving the performance while using only original video without enhancement as input during the inference phase.

According to our literature review, no existing studies fully addressed these three challenges concurrently. To address these challenges, we proposed a knowledge-distillation-based framework, Dual-Light KnowleDge Distillation for  Action Recognition in the Dark (DL-KDD). The DL-KDD framework overcome the challenges of feature learning from both original and enhanced video while avoiding the additional computational cost like two-stream methods do. Knowledge distillation \cite{kd}, in this context, serves as an effective method, helping the model to learn from the teacher model’s information without increasing the input feature set. 
Our architecture includes a teacher model consisting of an enhancement module and an action classifier to learn from enhanced features, and a student model which contains an action classifier to learn original features and teacher logit. Thus, during the inference phase, we only need to use the original video as input without any additional enhancement or dual input. To overcome the challenge of information completeness, we allow the student models to learn information from both light feature and dark videos, ensuring that all critical information is captured by the model. For the complexity tradeoff, we use knowledge distillation instead of a two-stream approach to allow the model to learn two types of features without including additional costs. For consistent performance, only original videos are required for the model during inference to achieve effective results.

As this is the first work that solves these three problems simultaneously, the main contributions are three-fold: 1) Our method utilizes both the original video and enhanced feature for action recognition in the dark. 2) Our model can use only original video input without additional features or enhancement during inference, thereby maintaining the model size while improving performance, and 3) We achieve state-of-the-art performance in dark human action recognition.

\section{Related Works}
\label{sec:headings}

\subsection{Action Recognition in the Dark}
In recent developments, various model architectures have been proposed for human action recognition, including those based on 3D-CNN  \cite{slowfast, 21d,3dres} and Transformers \cite{times,swin}. These technologies perform well under well-lit conditions. However, their performance degrades while facing low-light videos. As a result, innovative approaches \cite{darklight,gcn,dtcm} have been introduced to address the problem, where most of them selected 3D-CNN as the backbone due to their effectiveness. Chen et al. \cite{darklight} proposed DarkLight, which utilizes both original and frames enhanced by Gamma Intensity Correction (GIC) for action prediction. The method represents a significant advancement in this field, demonstrating the effectiveness of light enhancement for action recognition in the dark. The experiments also indicate that the dual-path approach, which utilizes both original and enhanced frames, captures more features than methods that estimate optical flow \cite{i3d}, thereby achieving better performance. Building on these advancements, recent studies \cite{gcn,dtcm} have further improved the accuracy of action prediction by incorporating ZeroDCE \cite{zerodce}, a light enhancement module. These studies have integrated ZeroDCE with backbone classifiers and showed remarkable performance gains. Consequently, the architecture that directly utilizes enhanced video as input has emerged as the most prevalent method in the field. Our DL-KDD addresses the issue of overlooking the importance of original video in recent studies, and the increased computational cost brought by two-stream methods by applying knowledge distillation with an enhancement module.

\subsection{Knowledge Distillation}
Knowledge distillation \cite{kd} has been applied across various sub-tasks within human action recognition, including cross-modality knowledge distillation \cite{cm1,cm2}, multi-view knowledge distillation \cite{mkdt}, and low-resolution action recognition \cite{resl}. Such methodologies improve model performance, particularly when limited input is available during inference. Lin and Tseng \cite{mkdt} proposed a Multi-view knowledge distillation framework that enables the model to efficiently learn from a single view while effectively capturing knowledge from all views. This demonstrates the capability of knowledge distillation to enable models to learn and integrate information from diverse sources. For light enhancement learning in dark environments, several studies \cite{lkd1, lkd2} have also utilized knowledge distillation and achieved notable success in the upstream enhancement task. 
Finally, for the specific task of human action recognition in the dark, Jin et al. \cite{ukd} highlighted the critical role of knowledge distillation, due to the high computational cost of video input. Their experiments showed that integrating knowledge distillation with optical flow and RGB features effectively supports model training. 
In our research, we aim to explore a novel approach by directly applying knowledge distillation to the downstream classification task using the enhanced feature. This strategy allows the student model to benefit from light enhancement while only requiring the original video input. By distilling the enhanced feature to the student model, our method can take advantage of light enhancement without the need for direct enhanced input, which optimizes both performance and computational cost.

\section{Proposed Methods}
\label{sec:others}
\begin{figure}[ht]
    \centering
    \includegraphics[width=\textwidth]{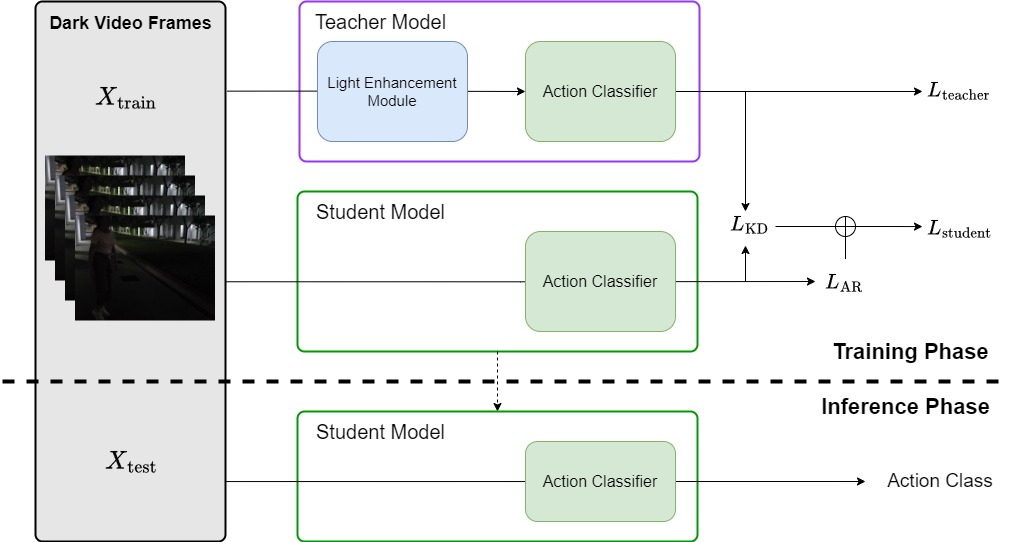}
    \caption{The overview architecture of our framework. The teacher model includes a light enhancement module and an action classifier. The student model includes only an action classifier. Knowledge distillation is applied to train the student model from the representation generated by the teacher model.}
    \label{fig:enter-label}
\end{figure}

\subsection{Problem Definition}
Action recognition aims to predict action labels from given input videos. Using the method of knowledge distillation, we train a teacher model to transfer its knowledge to a separate student model. Formally, let $D = \{(x_i, y_i)\}_{i=0}^n$ be a video-based dataset, where $x_i$ is a sample input and $y_i$ is the action class. During training phase, we train a teacher model $T$, which consists of an enhancement module $T_{e}$ and a backbone classifier $T_{c}$ , and a student model $S$ separately. For teacher model $T$, given training samples $X\{x_{1} ,x_{2},...,x_{j}\}$ from $D_{train}$ as the input of $T_{e}$, the module would generate enhanced samples $X'$. After that, $X'$ will be served as the input of $T_{c}$, which predict on $X'$ to the probability of each class, denote as $y^t$. The loss function is then applied on $y^t$ and the ground truth $y$.
\begin{equation}
y^t = T(x) = T_{c}(T_{e}(x))
\end{equation}
For student model $S$, given training samples $X\{x_{1} ,x_{2},...,x_{j}\}$ from $D_{train}$ as the input of $S$, the model generates the probability of each class, denoted as $y^s$. The loss function is then applied on $y^s$, $y^t$ and the ground truth $y$.
\begin{equation}
y^s = S(x)
\end{equation}
During the inference phase, only the student model $S$ will be in use for prediction.

\subsection{Overall Architecture}
The overall architecture is shown in Figure~\ref{fig:enter-label}. It consists of two main components: a teacher model and a student model. The teacher model includes a light enhancement module followed by an action classifier. The module first enhances the original video frames to improve visibility in order to extract features that may be loss in the dark. These enhanced features are then fed into the action classifier to generate feature representations.

After training the teacher model, we train the student model by taking original video frames without enhancement as input. This approach allows the student model to learn directly from the original data, which ensures that the model is not dependent on enhanced inputs during inference. The student model is an action classifier, and the training of the student model involves a dual learning process:

1. Direct Learning: The student model learns directly from the ground truth labels.

2. Distillation Learning: The student model is trained using the logits from the teacher model as soft targets, which allows the student model to learn from the feature representation that has been extracted from enhanced frames by teacher model. 

This architecture make use of both enhanced and original data, which optimizes the model’s performance without additional computational cost of using both data or processing enhancement during inference.

\subsection{Knowledge Distillation}
\textbf{Enhanced Feature Extraction}
The teacher model’s training begins with enhancing the original video frames. The enhancement module transforms the input $I$ into $I’$. This enhancement improve the feature visibility and information for action recognition in dark videos. After the enhancement, the enhanced frames $I’$ are fed into the action classifier for prediction. The action classifier processed the input frames into logits that capture the crucial information of the action label. A standard Cross-Entropy Loss is applied here for training the teacher model:
\begin{equation}
L_{\text{teacher}} = CrossEntropy(y, y^t)
\end{equation}
Where $y$ is the class label of the input video and $y^t$ is the logit extracted by the teacher model.

\textbf{Original Representation Learning}
Unlike the teacher model, the student model directly takes the original video frames $I$ as input. The action classifier is the same as that of the teacher models, but there is no enhancement module for the student model. The inputs are processed by the student model to generate the results $y^s$, and a cross-entropy loss is applied here to align the student model’s prediction with the ground truth.
\begin{equation}
L_{\text{AR}} = CrossEntropy(y, y^s)
\end{equation}

\textbf{Dual Knowledge Learning}
In addition to learning from the ground truth, a knowledge distillation process is applied to the student model, where it learns from the logit $y^t$ generated by the teacher model. This is achieved by minimizing the Kullback-Leibler divergence between the student model’s output $y^s$ and the soft targets provided by the teacher model, which indirectly transfers the enhanced feature knowledge from the teacher to the student model:
\begin{equation}
L_{\text{KD}} = KL(y^t, y^s)
\end{equation}

The overall loss function for training the student model is a combination of these two loss functions — Cross-Entropy loss for learning from the ground truth and KL divergence for knowledge transfer from the teacher model. The total loss function for the student model is denoted as:
\begin{equation}
L_{\text{student}}=\alpha L_{\text{AR}} + \beta L_{\text{KD}}
\end{equation}
where $\alpha$ and $\beta$ are the weights of the loss separately to balance the importance of the two training sources.

\section{Experiments}
\subsection{Experiment Settings}
\textbf{Dataset} We evaluated our method on three datasets: ARID, ARID V1.5 \cite{arid}, and Dark-48 \cite{dtcm}. The ARID \cite{arid} dataset has been a primary benchmark of dark human action recognition. It contains over 3780 video clips collected with 11 action classes. To further enhance the complexity of the dataset, ARID V1.5 was introduced. The video count is expanded to 5572, and the videos are collected from 24 scenes. The Dark-48 \cite{dtcm} dataset comprises 8815 dark videos from more than 40 scenes, featuring 48 classes with over 100 videos each. The dataset is split into training and testing sets in a ratio of 8:2. The training and testing settings in our experiment are the same as the original work\cite{arid, dtcm}. 

\textbf{Implementation Details} Inspired by \cite{darklight} , we selected R(2+1)D \cite{21d} followed by BERT \cite{bert} in replace of the conventional temporal global average pooling layer as our backbone classifier.  The backbone model was pre-trained on IG65M \cite{ig65} . For the enhancement module of the teacher model, we selected ZeroDCE \cite{zerodce} to generate enhanced frames. The input sequences were resized to 112 x 112 pixels, and the final input shape was 3 x 64 x 112 x 112 with batch size 2. We trained the teacher and the student model with AdamW \cite{adamw} optimizer with a learning rate 0.0001. The parameters for the loss function of the student model were set to 1 for both $\alpha$ and $\beta$.

\textbf{Metrics} In this task, we recorded top-1 and top-5 accuracy to evaluate the performance of the model. Since the ARID \cite{arid} dataset contains only 11 classes and most previous works have nearly reached 100\% top-5 accuracy, we primarily present the top-1 accuracy for ARID and ARID V1.5.

\subsection{Ablation Study}
In this section, we focus on an ablative comparison of the ARID dataset to demonstrate the effectiveness of our proposed framework. To illustrate the efficacy of our training method, we present results comparing the backbone model trained with and without our method. Additionally, comparisons between the teacher and student models are displayed to show that the student network can achieve better results even without enhancement after the knowledge distillation training. Table~\ref{tab:ablation} provides a detailed display of the final performance of the teacher, student model of DL-KDD, and the performance of similar architecture without knowledge distillation training method.
\begin{table}[ht]
\centering
\caption{Comparative Performance of DL-KDD on ARID}
\label{tab:ablation}
\begin{tabular}{lc}
\hline
\textbf{Model} & \textbf{Top-1 Accuracy (\%)} \\
\hline
R(2+1)D + BERT \cite{darklight} & 92.44 \\
\hline
DL-KDD-Teacher: ZeroDCE + R(2+1)D + BERT & 95.73 \\
\textbf{DL-KDD-Student: R(2+1)D + BERT (ours)} & \textbf{97.27} \\
\hline
\end{tabular}
\end{table}

\textbf{With and without Knowledge Distillation} As shown in Table~\ref{tab:ablation}, our knowledge distillation training method improved the performance of the same architecture by 4.83\%, which shows the effectiveness of learning from the knowledge distilled from the enhanced feature by the teacher model. With the additional knowledge provided by the teacher model, the student model can take advantage of enhanced representation even without enhanced feature inputs during testing.

\textbf{Comparison with Teacher Model} The comparison between the student and teacher model shows that even the student model uses a simpler architecture without enhancement, it achieves an improvement of 1.54\% over the teacher model, which indicates that in addition to the distilled knowledge of enhanced features, the original video also contains critical information that improves model performance. By learning from original inputs, the student model accesses additional information from enhanced features, resulting in better performance than the teacher model.

\subsection{Comparison with State-of-the-Art}
We conduct extensive experiments to compare our work with the recent state-of-the-art methods in dark human action recognition, including DarkLight \cite{darklight}, DTCM \cite{dtcm}, and R(2+1)D-GCN+BERT \cite{gcn} across the ARID, ARID V1.5 \cite{arid}, and Dark48 \cite{dtcm} dataset. Partial results from previous works are collected from \cite{darklight,gcn,dtcm}. Table~\ref{tab:arid_results} and Table~\ref{tab:arid_v1_5_results} present detailed results for the two versions of the ARID dataset. Despite high baseline performances on these datasets, our proposed method outperforms existing models and achieves the best results. Table~\ref{tab:dark_48_results} indicates that our model reached a Top-1 accuracy of 50.86\% on the Dark-48 dataset. This demonstrates a significant improvement over the best previously reported result on Dark-48 by 4.18\%. These results illustrate that our proposed knowledge distillation framework successfully enhances the information learned by the model, which enables our model to achieve the best performance while using only the original video input during testing.

\begin{table}[ht]
\centering
\begin{minipage}{.48\linewidth}
\centering
\caption{Results comparison on ARID}
\label{tab:arid_results}
\begin{tabular}{|l|c|}
\hline
\textbf{Model} & \textbf{Top-1 Accuracy (\%)} \\
\hline
I3D-RGB & 68.29 \\
I3D Two-stream & 72.78 \\
3D-ResNext-101 & 74.73 \\
DarkLight & 94.04 \\
DTCM & 96.36 \\
R(2+1)D-GCN+BERT & 96.60 \\
\hline
\textbf{DL-KDD (Ours)} & \textbf{97.27} \\
\hline
\end{tabular}
\end{minipage}\hfill
\begin{minipage}{.48\linewidth}
\centering
\caption{Results comparison on ARID V1.5}
\label{tab:arid_v1_5_results}
\begin{tabular}{|l|c|}
\hline
\textbf{Model} & \textbf{Top-1 Accuracy (\%)} \\
\hline
I3D-RGB & 48.75 \\
I3D Two-stream & 51.24 \\
DarkLight & 84.13 \\
R(2+1)D-GCN+BERT & 86.93 \\
\hline
\textbf{DL-KDD (Ours)} & \textbf{88.04} \\
\hline
\end{tabular}
\end{minipage}
\end{table}

\begin{table}[ht]
\centering
\caption{Results comparison on Dark-48}
\label{tab:dark_48_results}
\begin{tabular}{|l|c|c|}
\hline
\textbf{Model} & \textbf{Top-1 Accuracy (\%)} & \textbf{Top-5 Accuracy (\%)} \\
\hline
I3D-RGB & 32.25 & 65.35 \\
3D-ResNext-101 & 37.23 & 68.86 \\
DarkLight & 42.27 & 70.47 \\
DTCM & 46.68 & 75.92 \\
\hline
\textbf{DL-KDD (Ours)} & \textbf{50.86} & \textbf{78.29} \\
\hline
\end{tabular}
\end{table}

\newpage
\section{Conclusion}
In this work, we have proposed a novel knowledge-distillation-based framework named DL-KDD for dark human action recognition, emphasizing the importance of utilizing both the original video and the enhanced feature to prevent the loss of original information. Moreover, the proposed framework avoids the additional cost brought by two-stream methods. We effectively distill the knowledge of light enhancement to the student model, enabling the student model to use only original videos as input during inference and achieve better results. The state-of-the-art performance on the ARID and Dark-48 datasets proves the effectiveness of our method. For future work, we will continue to refine our architecture for further advancement on dark human action recognition.

%Bibliography
\bibliographystyle{unsrt}  
\bibliography{references}

\end{document}